\documentclass[10pt,twocolumn,letterpaper]{article}

\usepackage{iccv}
\usepackage{times}
\usepackage{epsfig}
\usepackage{graphicx}
\usepackage{amsmath}
\usepackage{amssymb}
\usepackage{subfigure}
\usepackage{url}
\graphicspath{{pic/}}

 \usepackage[pagebackref=true,breaklinks=true,letterpaper=true,colorlinks,bookmarks=false]{hyperref}

\usepackage[breaklinks=true,bookmarks=false]{hyperref}

\iccvfinalcopy 

\def\iccvPaperID{****} 

\ificcvfinal\fi

\begin{document}


\title{Mask-Guided Attention Network for Occluded Pedestrian Detection}

\author{Yanwei Pang$^1$,~Jin Xie$^{1}$,~Muhammad Haris Khan$^2$,~Rao Muhammad Anwer$^2$,\\~Fahad Shahbaz Khan$^{2,3}$,~Ling Shao$^2$\\
	{\normalsize $^1$Tianjin University~~~~$^2$Inception Institute of Artificial Intelligence, UAE~~~~$^3$CVL, Link\"oping University, Sweden}\\
	{\tt\small \{pyw,jinxie\}@tju.edu.cn, \{muhammad.haris,rao.anwer,fahad.khan,ling.shao\}@inceptioniai.org}
}
\maketitle
\ificcvfinal\thispagestyle{empty}\fi

\begin{abstract}

Pedestrian detection relying on deep convolution neural networks has made significant progress. Though promising results have been achieved on standard pedestrians, the performance on heavily occluded pedestrians remains far from satisfactory. The main culprits are intra-class occlusions involving other pedestrians and inter-class occlusions caused by other objects, such as cars and bicycles. These result in a multitude of occlusion patterns. We propose an approach for occluded pedestrian detection with the following contributions. First, we introduce a novel mask-guided attention network that fits naturally into popular pedestrian detection pipelines. Our attention network emphasizes on visible pedestrian regions while suppressing the occluded ones by modulating full body features. Second, we empirically demonstrate that coarse-level segmentation annotations provide reasonable approximation to their dense pixel-wise counterparts. Experiments are performed on CityPersons and Caltech datasets. Our approach sets a new state-of-the-art on both datasets. Our approach obtains an absolute gain of 9.5\% in log-average miss rate, compared to the best reported results \cite{citypersons_2017_zhang} on the heavily occluded \textbf{HO} pedestrian set of CityPersons test set. Further, on the \textbf{HO} pedestrian set of Caltech dataset, our method achieves an absolute gain of 5.0\% in log-average miss rate, compared to the best reported results \cite{Lin_2018_ECCV}. Code and models are available at: \url{https://github.com/Leotju/MGAN}.

\end{abstract}
\section{Introduction}
\label{sec:introduction}
Pedestrian detection is a challenging computer vision problem with numerous real-world applications. Recently, deep convolutional neural networks (CNNs) have pervaded many areas of computer vision ranging from object recognition \cite{simonyan2014vgg, He_2016_CVPR, GlanceNets, CIC}, to generic object detection \cite{fasterrcnn_2015_nips,liu2016ssd, lin2018focal}, to pedestrian detection \cite{JimmyRenCVPR17, WeiLiuECCV18, ZhaoweiCaiECCV16, JiayuanMaoCVPR17, GarrickBrazilICCV17, XinlongWangCVPR18}.

\begin{figure}[t]
   \centering
	\resizebox{\linewidth}{!}{
	\begin{tabular}{ccc}
		  \includegraphics[height=3cm]{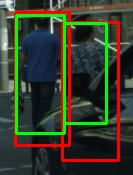}& 
	     \includegraphics[height=3cm]{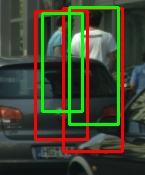}&
	     \includegraphics[height=3cm]{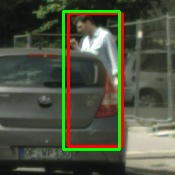}\\
	     \includegraphics[height=3cm]{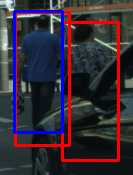}& 
	     \includegraphics[height=3cm]{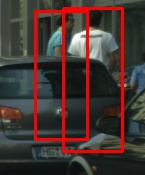}&
	     \includegraphics[height=3cm]{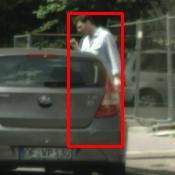}\\
	     \multicolumn{3}{c}{\includegraphics[width=0.95\linewidth]{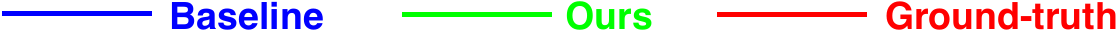}}
	\end{tabular}
	
	}
   
     \caption{Detection examples using our approach (top row) and the baseline Faster R-CNN \cite{fasterrcnn_2015_nips} (bottom row). For improved visualization, detection regions are cropped from images of CityPersons val. set \cite{citypersons_2017_zhang}. All results are obtained using the same false positive per image (FPPI) criterion. Our approach robustly handles 
     occlusions, yielding higher recall for occluded pedestrians. }
    \label{fig:comparison_baseline_MGAN} \vspace{-0.3cm}
\end{figure}

Despite the recent progress on standard benchmarks with non-occluded or reasonably occluded pedestrians, state-of-the-art approaches still struggle under severe occlusions. For example, when walking in close proximity, a pedestrian is likely to be obstructed by other pedestrians and/or other objects like cars and bicycles. For illustration, Fig.~\ref{fig:comparison_baseline_MGAN} displays the performance of baseline Faster R-CNN pedestrian detector \cite{fasterrcnn_2015_nips} under heavy occlusions. Handling occlusions is a key challenge; they occur frequently in real-world applications of pedestrian detection. Therefore, recent benchmarks specifically focus on heavily occluded pedestrian detection. For instance, CityPersons \cite{citypersons_2017_zhang} dataset has around 70$\%$ of pedestrians depicting various degrees of occlusions. 

Most existing approaches employ a holistic detection strategy \cite{JimmyRenCVPR17, WeiLiuECCV18, ZhaoweiCaiECCV16, JiayuanMaoCVPR17} that assumes entirely visible pedestrians when trained using full body annotations. However, such a strategy is sub-optimal under partial or heavy occlusions since most of the pedestrian's body is invisible. This deteriorates the performance by degrading the discriminative ability of the pedestrian model due to the inclusion of background regions inside the full body detection window.

Lately, several pedestrian detection methods \cite{WanliOuyangICCV13,MarkusMathiasICCV13,YonglongTianICCV15,ChunluanZhouICCV17, cascade} tackle occlusions by learning a series of part detectors that are integrated to detect partially occluded pedestrians. They either learn an ensemble model and integrate their outputs or jointly train different occlusion patterns to handle occlusions. Ensemble-based approaches are computationally expensive which prohibits real-time detection. On the other hand, methods based on joint learning of occlusion patterns are difficult to train and rely on fusion of part detection scores. Instead, we investigate occluded pedestrian detection without explicitly using part information.

In contrast to part-based approaches for handling occlusions, a few methods \cite{ShanshanCVPR18, Zhou_2018_ECCV} exploit visible-region information, available with standard pedestrian detection benchmarks \cite{citypersons_2017_zhang,Dollar_2012_PAMI}, to either output visible part regions for proposal generation \cite{Zhou_2018_ECCV} or employ as extraneous supervision to learn occlusion patterns \cite{ShanshanCVPR18}. In this work, we follow the footsteps of these recent methods to tackle the problem of occluded detection. Different to \cite{Zhou_2018_ECCV, ShanshanCVPR18}, we make use of visible body information to produce a pixel-wise spatial attention to modulate the multichannel features in the standard full body estimation branch. The proposed mask-guided spatial attention network can be easily integrated into mainstream pedestrian detectors and is not limited to specific occlusion patterns. Fig. \ref{fig:comparison_baseline_MGAN} shows that the proposed approach is able to detect occluded pedestrians over a wide spectrum ranging from partial to heavy occlusions.   

\noindent\textbf{Contributions:} We propose a deep architecture termed as Mask-Guided Attention Network (MGAN), which comprises two branches: the Standard Pedestrian Detection branch and a novel Mask-Guided Attention branch. The Standard Pedestrian Detection branch generates features using full body annotations for supervision. The proposed Mask-Guided Attention Branch produces a pixel-wise attention map using visible-region information, thereby highlighting the visible body region while suppressing the occluded part of the pedestrian. The spatial attention map is then deployed to modulate the standard full body features by emphasizing regions likely belonging to visible part of the pedestrian. Further, we empirically demonstrate that for occluded pedestrian detection, the weak approximation of dense pixel-wise annotations yields similar results.

We perform experiments on two pedestrian detection benchmarks: CityPersons \cite{citypersons_2017_zhang} and Caltech \cite{Dollar_2012_PAMI}. On both datasets, our approach displays superior results compared to existing pedestrian detection methods. Further, our approach improves the state-of-the-art \cite{Zhou_2018_ECCV} from $44.2$ to $39.4$ in log-average miss rate on the \textbf{HO} set of CityPersons, which has 35-80$\%$ occluded pedestrians, using the \textit{same} level of supervision, input scale and backbone network.

\section{Related Work}

\textbf{Deep Pedestrian Detection.} Recently, pedestrian detection approaches based on deep learning techniques have exhibited state-of-the-art performance \cite{JimmyRenCVPR17, WeiLiuECCV18, ZhaoweiCaiECCV16, JiayuanMaoCVPR17, GarrickBrazilICCV17, XinlongWangCVPR18, zhang2016faster, du2017fused}. CNN-based detectors can be roughly divided into two categories: the two-stage approach comprising separate proposal generation followed by confidence computation of proposals and the one-stage approach where proposal generation and classification are formulated as a single-stage regression problem. 
Most existing pedestrian detection methods either employ the single-stage \cite{JimmyRenCVPR17, WeiLiuECCV18, JunhyugNohCVPR18} or two-stage strategy \cite{ZhaoweiCaiECCV16, JiayuanMaoCVPR17, GarrickBrazilICCV17, XinlongWangCVPR18} as their backbone architecture. The work of \cite{JimmyRenCVPR17} proposed a recurrent rolling convolution architecture that aggregates useful contextual information among the feature maps to improve single-stage detectors. Liu \etal \cite{WeiLiuECCV18} extended the single-stage architecture with an asymptotic localization fitting module storing multiple predictors to evolve default anchor boxes. This improves the quality of positive samples while enables hard negative mining with increased thresholds. 

In the two-stage detection strategy, the work of \cite{ZhaoweiCaiECCV16} proposed a deep multi-scale detection approach where intermediate network layers, with receptive fields similar to different object scales, are employed to perform the detection task. Mao \etal \cite{JiayuanMaoCVPR17} proposed to integrate channel features (\ie, edge, heatmap, optical flow and disparity) into a two-stage deep pedestrian detector. The work of \cite{GarrickBrazilICCV17} introduced a multi-task approach for joint supervision of pedestrian detection and semantic segmentation. The segmentation infusion layer is employed to highlight pedestrians, thereby enabling downstream detection easier. The work of \cite{Di_Cheneccv18} employed a two-stage pre-trained person detector (Faster R-CNN) and an instance segmentation model for person re-identification. Each detected person is cropped out from the original image and fed to another network. Wang \etal \cite{XinlongWangCVPR18} introduced repulsion losses that prevent a predicted bounding-box from shifting to neighbouring overlapped objects to counter occlusions. Due to their superior performance on pedestrian benchmarks \cite{citypersons_2017_zhang}, we deploy two-stage detection strategy as backbone pipeline in our work.

\textbf{Occlusion Handling in Pedestrian Detection.} Several works investigated the problem of handling occlusions in pedestrian detection. A common strategy \cite{MarkusMathiasICCV13, zhou2014non, WanliOuyangICCV13, YonglongTianICCV15, ChunluanZhouICCV17} is the part-based approach where a set of part detectors are learned with each part designed to handle a specific occlusion pattern. Some of these part-based approaches \cite{MarkusMathiasICCV13,YonglongTianICCV15} train an ensemble model for most occurring occlusion patterns and are computationally expensive due to the deployment of large number of part detectors. Alternatively, some part-based approaches \cite{WanliOuyangICCV13, ChunluanZhouICCV17} rely on joint learning of collection of parts to capture occlusion patterns. 

\begin{figure*}[th]
    \centering
     \includegraphics[width=1.0\linewidth]{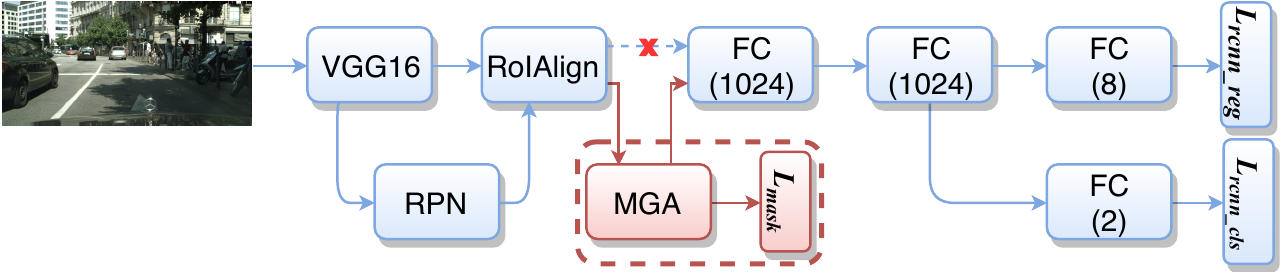}
    \caption{The overall network architecture of our Mask-Guided Attention Network (MGAN). It comprises a standard pedestrian detection (SPD) branch, whose components are shown in blue. It introduces a novel Mask-Guided Attention (MGA) module enclosed in red dashed box. Note, after RoI Align there is a classification stage in the SPD branch whose first layer is shown by FC (1024). In our architecture, standard full body features in SPD branch after RoI Align layer are modulated by MGA branch before getting scored by the classification stage. This is in contrast to baseline SPD where these features directly become the input to the classification stage without any modulation. 
     }\vspace{-0.3cm}
    \label{fig:mgfe-net}
\end{figure*}

Contrary to the aforementioned methods, recent approaches have exploited visible body information either as an explicit branch to regress visible part regions for proposal generation \cite{Zhou_2018_ECCV} or as external guidance to learn specific  occlusion modes (full, upper-body, left-body and right-body visible) in a supervised fashion \cite{ShanshanCVPR18}. Different to \cite{Zhou_2018_ECCV}, we utilize the visible branch to generate a pixel-wise attention map that is used to modulate multi-channel convolutional features in the standard full body estimation branch. Unlike ATT-vbb \cite{ShanshanCVPR18}, we propose a spatial attention network that is not restricted to only certain type of occlusion patterns. Further, when using the \textit{same} level of supervision, input scale, backbone and training data our approach provides a significant gain of 4.8\% and 5.6\% compared to \cite{Zhou_2018_ECCV} and \cite{ShanshanCVPR18}, respectively on \textbf{HO} set of CityPersons.

\section{Proposed Approach}
We propose Mask-Guided Attention Network (MGAN) that features a novel Mask-Guided Attention branch. It produces a pixel-wise attention map, highlighting the visible body part while suppressing the occluded part in the full body features. This branch is a lightweight, easy to implement module and is integrated into the standard pedestrian pipeline, thereby making a single, coherent architecture capable of end-to-end training.

The overall proposed architecture comprises two main branches: a Standard Pedestrian Detector (SPD) branch that detects pedestrian \cite{fasterrcnn_2015_nips} using full body information whom components are shown in blue in Fig.~\ref{fig:mgfe-net}, and a novel Mask-Guided Attention (MGA) branch that produces a pixel-wise attention map employing visible bounding-box information. This branch modulates the full body features and shown with a red dashed box in Fig.~\ref{fig:mgfe-net}. Next, we review the SPD branch and then detail the design of our MGA branch. 

\subsection{Standard Pedestrian Detector Branch}
\label{subsection:modulation branch}
We choose Faster R-CNN\cite{fasterrcnn_2015_nips} as the standard pedestrian detection branch mainly for its state-of-the-art performance. It takes a raw image as input, first deploys a pre-trained ImageNet model such as VGG-16 \cite{simonyan2014vgg} and then a region proposal network (RPN) to generate region proposals. Extracts proposal features by cropping the corresponding region-of-interest (RoI) in the extracted feature maps and further resizes them to fixed dimensions with a RoI pooling layer. Note, we replace RoI pooling layer with RoI Align layer \cite{maskrcnn_2017_iccv_he} in our experiments. This makes every proposal to have same feature length. These features go through a classification net that generates the classification score (\ie the probability that this proposal contains a pedestrian) and the regressed bounding box coordinates for every proposal. Fig.~\ref{fig:mgfe-net} visually illustrates the aforementioned steps. Since every layer in Faster R-CNN is differentiable, it is trainable end-to-end with the following loss function:

\begin{equation}
	{L}_{0} = {L}_{rpn}+{L}_{rcnn}.
    \label{loss:faster_rcnn}
\end{equation}
Each term has a classification loss and a bounding box regression loss. Thus, Eq. \ref{loss:faster_rcnn} can be written as:
\begin{equation}
	{L}_{0} = {L}_{rpn\_cls}+{L}_{rpn\_reg} + {L}_{rcnn\_cls}+{L}_{rcnn\_reg},
    \label{loss:faster_rcnn2}
\end{equation}
where ${L}_{rpn\_cls}$ and ${L}_{rcnn\_cls}$ refer to the classification loss of RPN and R-CNN, respectively, and ${L}_{rpn\_reg}$ and ${L}_{rcnn\_reg}$ are the bounding box regression loss of RPN and R-CNN, respectively. Here, classification loss is Cross-Entropy loss and the bounding box regression loss is Smooth-L1 loss.

\begin{figure}
     \centering
        \includegraphics[width=1.0\linewidth]{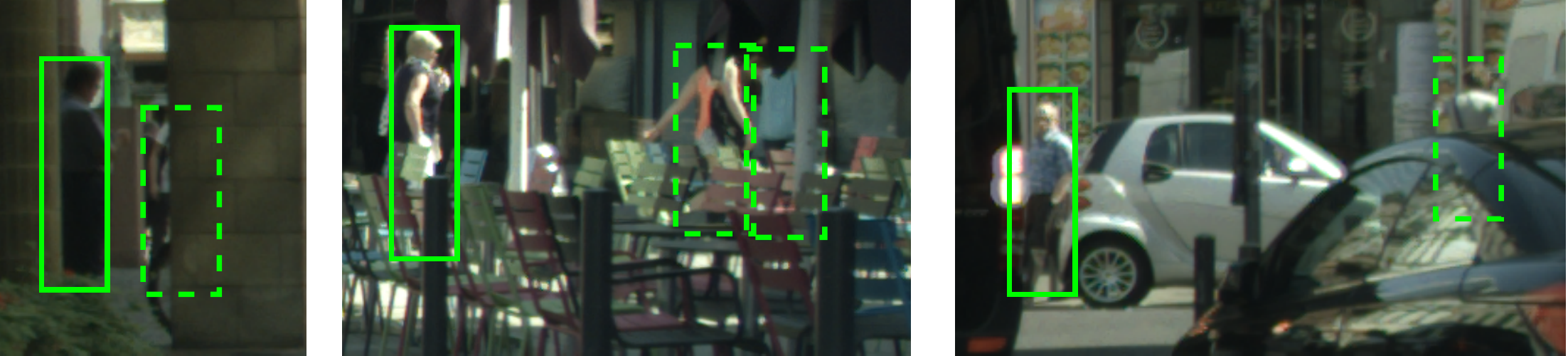}
     \caption{Results of a pedestrian detector trained by full body bounding-box annotations. We show three different occluded scenarios. Solid green boxes denote predictions by the detector and dashed green boxes represent the missed detections. The detector cannot capture heavily occluded pedestrians and might result in high miss rate under similar circumstances.} 
     \label{fig:re-bbox-anno}\vspace{-0.3cm}
 \end{figure}

\textbf{Discussion.} Despite achieving impressive results for non-occluded pedestrians, this and similar pipelines struggle - showing high miss rates - in the presence of partial and heavy occlusions. Fig.~\ref{fig:re-bbox-anno} depicts pedestrian detector trained using full body bounding-box annotations produces less false positives but miss several pedestrians. This is likely due to the contribution of features towards the scoring of a proposal corresponding to the occluded parts of the pedestrian. As the occlusion modifies the pedestrian appearance, the features for the occluding part are vastly different to the visible part. We show how to suppress these (occluded) features and enhance the visible ones to obtain more robust features~\footnote{One might argue that a simple solution can be to train a pedestrian detector supervised only by visible-region annotations. Though the resulting detector will capture occluded pedestrians and will decrease the miss rate, it would result in high false positive detections.}. We present a \textbf{mask-guided spatial attention} approach that greatly alleviates the impact of occluded features while stresses the visible-region features, and is not restricted to certain occlusion types. This mask-guided attention network is a lightweight CNN branch integrated into the standard pedestrian detection network.

   \vspace{0.2cm}
\subsection{Mask-Guided Attention Branch}
\begin{figure}
    \centering
    \subfigure[]{
    \includegraphics[height=3cm]{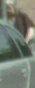}}
    \subfigure[]{
    \includegraphics[height=3cm]{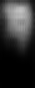}}
    \subfigure[]{
    \includegraphics[height=3cm]{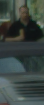}}
    \subfigure[]{
    \includegraphics[height=3cm]{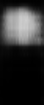}}
    \subfigure[]{
    \includegraphics[height=3cm]{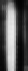}}
    \subfigure[]{
    \includegraphics[height=3cm]{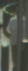}}
    \caption{Spatial Attention Masks generated by our MGA branch. Three spatial attention masks correspond to differently occluded pedestrians \ie partial and heavy. Note the enhancing of visible part and the hiding of occluded part in each mask.}
    \label{fig:visualization} \vspace{-0.1cm}
\end{figure}

The proposed mask-guided attention branch is highlighted with red annotated box in Fig.~\ref{fig:mgfe-net}. It produces a spatial attention mask supervised by visible-region bounding box information and using this modulates the multichannel features generated by the RoI Align layer. Fig.~\ref{fig:visualization} shows three different occluded persons and their corresponding spatial attention masks. These masks accurately reveal the visible part and hide the occluded part for three variable occlusion patterns. Modulated features with these masks help classification network detect partially and heavily occluded pedestrians with higher confidence, which otherwise might not get detected due to being scored poorly. The following subsections detail our mask-guided attention branch.

 \vspace{-0.1cm}
\subsubsection{MGA Architecture}

\begin{figure}
    \centering
    \includegraphics[width=1.0\linewidth]{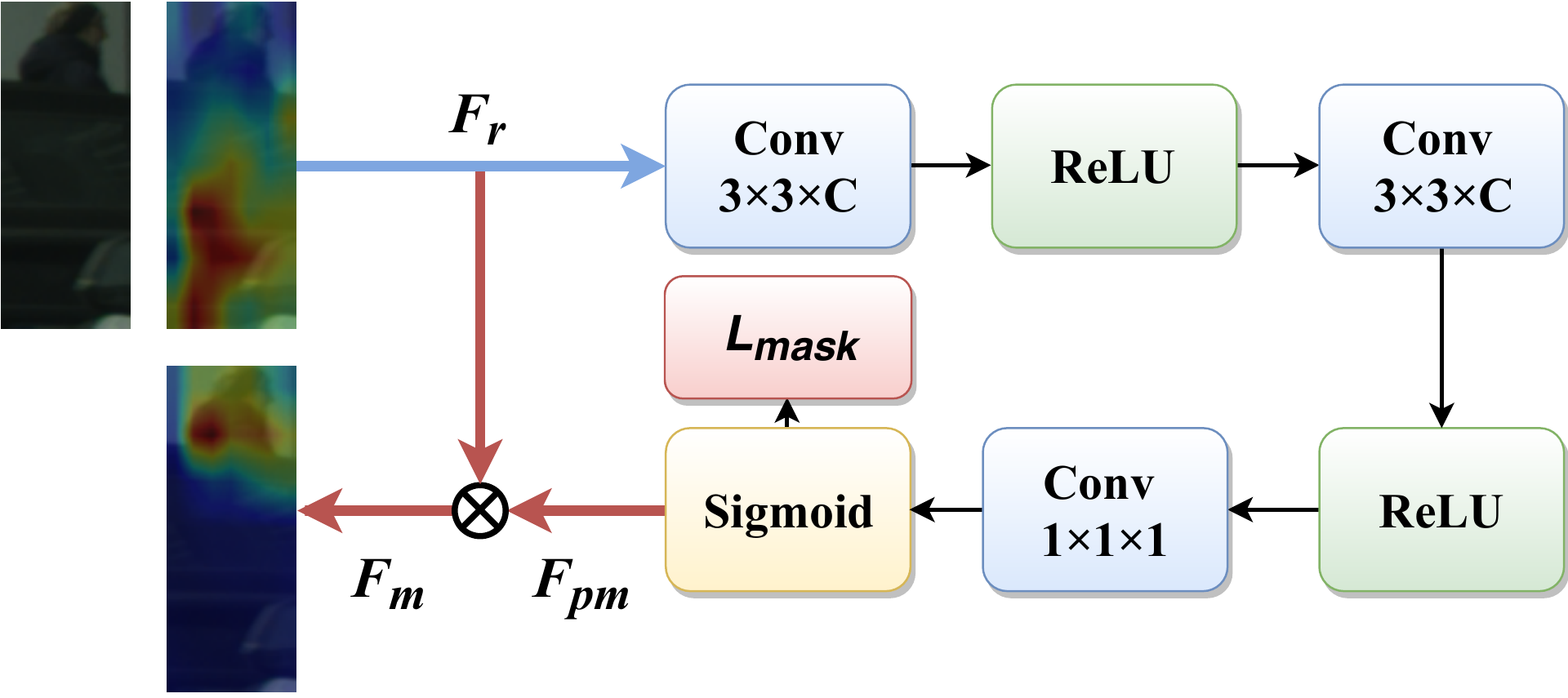}
    \caption{The network architecture of our Mask-Guided Attention (MGA) Branch. It takes RoI features and generates modulated features using a small stack of conv. operations, followed by ReLU nonlinearities.}
    \label{fig:mgfe-detail} \vspace{-0.2cm}
\end{figure} 

The proposed MGA branch architecture is depicted in Fig.~\ref{fig:mgfe-detail}. The input to MGA branch are the multichannel features from RoI Align layer and the output are the modulated multichannel features. The modulated features are generated using pedestrian probability map, termed as the spatial attention mask.
We denote input features as ${F}_{r} \in [H\times W\times C]$, where first two dimensions are the resolution and the last one is the depth. Firstly, two $3\times3$ filter size convolution layer followed by Rectified Linear Unit (ReLU) extracts features. 
 
Then, a $1\times1$ filter size conv. layer followed by a sigmoid layer generates the probability map ${F}_{pm} \in [H\times W \times1]$. 
In our experiments, $H$ and $W$ are set to 7, and $C$ is set to 512.

\begin{figure}
    \centering
    \subfigure[]{\includegraphics[height=3cm]{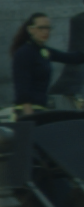}}
    \subfigure[]{\includegraphics[height=3cm]{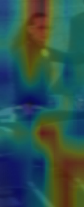}}
    \subfigure[]{\includegraphics[height=3cm]{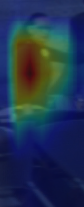}}
    \subfigure[]{\includegraphics[height=3cm]{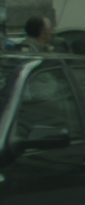}}
    \subfigure[]{\includegraphics[height=3cm]{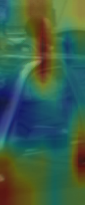}}
    \subfigure[]{\includegraphics[height=3cm]{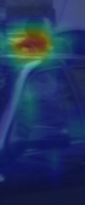}}
    
    \caption{Visual comparison of RoI features and corresponding modulated features. (a) and (d) are two different proposals. (b) and (e) depict their corresponding RoI features. (c) and (f) show their corresponding modulated features. In contrast to RoI features, modulated features from our MGA branch have visible region signified and occluded part concealed.}
    \label{fig:comparison_original_modulated} \vspace{-0.4cm}
\end{figure}

These probability maps ${F}_{pm}$ modulate the multichannel features ${F}_{r}$ of a proposal to obtain re-weighted features ${F}_{m}$. We achieve this by taking the element-wise product of every feature channel in ${F}_{r}$ with ${F}_{pm}$ as:
\begin{equation}
{F}_{{m}_{i}} ={F}_{{r}_{i}} \odot {F}_{pm}, i=1,2,...,C, 
\end{equation}
where $i$ is the channel index and $\odot$ is the element-wise product. Instead of RoI features ${F}_{r}$, we feed modulated features ${F}_{m}$ to the classification net for scoring proposals. Fig.~\ref{fig:comparison_original_modulated} illustrates that in contrast to RoI features, modulated features from MGA branch have visible region signified and occluded part concealed thereby leading to a relatively high confidence for occluded proposals.   

 \vspace{-0.1cm}
\subsubsection{Coarse-level Segmentation Annotation}
\label{subsubsection:Coarse-level Segmentation Annotation}

The spatial attention mask for a proposal and image-level segmentation requires supervision in the form of dense pixel-wise segmentation annotation. This, however, is tedious to acquire in many computer vision tasks including pedestrian detection. We therefore adapt visible-region bounding box annotation as an approximate alternative. Such annotations are readily available for the popular pedestrian detection benchmarks \cite{citypersons_2017_zhang, Dollar_2012_PAMI}.

The adaption is as follows. If a pixel lies in the visible-region bounding-box annotation; it is a foreground pixel with a label one. Similarly, a pixel outside this region is a background pixel and its label is zero. This labelling process creates a coarse-level segmentation annotation. Importantly, such weakly labelled annotations have generated accurate masks in our experiments (see Fig.~\ref{fig:visualization}). Description
of MGA branch finishes here and the following subsection
discusses the loss function optimized in proposed approach.

 \vspace{-0.1cm}
\subsubsection{Loss Function}
Here, we present our loss function for the proposed architecture MGAN. The overall loss formulation $L$  is:   

\begin{equation}
{L}={{L}_{0}}+ \alpha{{L}_{mask}} + \beta L_{occ},
\end{equation}
where ${L}_{0}$ is the loss for Faster R-CNN as in Eq.(\ref{loss:faster_rcnn}), ${L}_{mask}$ is the loss term for the proposed MGA branch, and $L_{occ}$ is the occlusion-sensitive loss term. Note that we tend to jointly optimize all the losses in the spirit of end-to-end training. In our experiments, we set $\alpha = 0.5, \beta=1$  by default. ${L}_{mask}$ and $L_{occ}$ are defined on positive proposals. ${L}_{mask}$ on coarse-level (weak) supervision is formulated as a per-pixel binary cross-entropy loss (BCE loss):

\begin{equation}
    {L}_{mask} = BCELoss({p}_{n}(x,y), \hat{p}_{n}(x,y)),
\end{equation}

where ${\hat{p}}_{n}(x,y)$ are the predictions produced by MGA branch and ${{p}}_{n}(x,y)$ represents the ground truth \ie, coarse-level segmentation annotations.

Further, to make the classification loss aware of variable occlusion levels, we introduce an occlusion sensitive loss term $L_{occ}$. It simply weights pedestrian training proposals based on their occlusion levels, derived from ${{p}}_{n}(x,y)$, when computing the standard cross-entropy loss (CE loss):

\begin{equation}
    \begin{split}
L_{occ} = \frac{1}{N} \sum_{n=1}^{N}& \{  [1 - \frac{1}{WH}\sum_x^W \sum_y^H {{p}}_{n}(x,y)]\\
    &CELoss({p}_{n}^{rcnn\_cls}, \hat{p}_{n}^{rcnn\_cls}) \},
\end{split}
\end{equation}
 
where $W$ and $H$ are the width and the height of pedestrian probability map. $\hat{p}_{n}^{rcnn\_cls}$ are the predictions produced by the classification branch of RCNN, and ${p}_{n}^{rcnn\_cls}$ represents the ground-truth.

\section{Experiments}
\subsection{Datasets and Evaluation Metrics}
\textbf{Datasets.} We perform experiments on two pedestrian detection benchmarks: CityPersons \cite{citypersons_2017_zhang} and Caltech \cite{Dollar_2012_PAMI}. CityPersons \cite{citypersons_2017_zhang} is a challenging dataset for pedestrian detection and exhibits large diversity. It consists of 2975 training images, 500 validation images, and 1575 test images. Caltech pedestrian is a popular dataset \cite{Dollar_2012_PAMI} featuring 11 sets of videos. First 6 sets (0-5) correspond to training and the last 5 sets (6-10) are for testing. To increase training set size, the frames are sampled at 10Hz. The test images are captured at 1 Hz. Finally, the training and test sets have 42782 and 4024 images, respectively. Both datasets provide box annotations for full body and visible region.\\
\textbf{Evaluation Metrics.}
We report performance using standard average-log miss rate (MR) in experiments; it is computed over the false positive per image (FPPI) range of $[{10}^{-2},{10}^{0}]$ \cite{Dollar_2012_PAMI}.  We select ${MR}^{-2}$ and its lower value reflects better detection performance. 
On the Caltech dataset, we report results across three different occlusion degrees: Reasonable (\textbf{R}), Heavy (\textbf{HO}) and the combined Reasonable + Heavy (\textbf{R+HO}). For the CityPersons dataset, we follow \cite{citypersons_2017_zhang} and report results on Reasonable (\textbf{R}) and Heavy (\textbf{HO}) sets. The visibility ratio in \textbf{R} set is larger than $65 \%$, and the visibility ratio in \textbf{HO} set ranges from $20 \%$ to $65 \%$. Similarly, the visibility ratio in \textbf{R + HO} set is larger than $20 \%$. In all subsets, the height of pedestrians over 50 pixels is taken for evaluation, as in \cite{ShanshanCVPR18}. Note that the \textbf{HO} set is designed to evaluate performance in case of severe occlusions. 

\subsection{Implementation and Training Details}
For both datasets, the networks are trained on a NVIDIA GPU and a mini-batch comprises 2 image per GPU. We select the Adam \cite{kingma2014adam} solver as optimizer.
We now detail settings specific to the two datasets. \\
\textbf{CityPersons.} We fine-tune the ImageNet pretrained VGG-16 \cite{simonyan2014vgg} models on CityPersons trainset. 
Except we use two fully-connected layers with 1024 output dimensions instead of 4096 output dimensions, we follow the same experimental protocol as in \cite{citypersons_2017_zhang}. 
We start with the initial learning rate of $1\times{10}^{-4}$ for the first $8$ epochs and further decay it to $1\times{10}^{-5}$ and perform $3$ epochs. \\
\textbf{Caltech.} We start with the model pretrained on CityPersons dataset. To fine-tune the model, an initial learning rate of ${10}^{-4}$ is used for first $3$ training epochs. The training is further performed for another $1$ epoch after decaying the initial learning rate by a factor of 10.

\subsection{Ablation Study on CityPersons Dataset} 
We evaluate our approach (MGAN) by performing an ablation study on CityPersons dataset. 

\noindent\textbf{Baseline Comparison.} Tab.~\ref{tab:baseline_comparison} shows the baseline comparison. For a fair comparison, we use the same set of ground-truth pedestrian examples during training for all methods. We select ground-truth pedestrian examples which are at least 50 pixels tall with visibility $\ge65\%$ for the training purpose. The baseline SPD detector obtains a log-average miss rate of $13.8\%$ and $57.0\%$ on \textbf{R}  and \textbf{HO} sets of CityPersons dataset, respectively. Our Final MGAN based on the MGA branch and occlusion-sensitive loss term significantly reduces the error on both \textbf{R}  and \textbf{HO} sets. Under heavy occlusions (\textbf{HO}), our MGAN achieves an absolute reduction of $5.3\%$ in log-average miss rate, compared to the baseline. The significant reduction in error on the (\textbf{HO}) set demonstrates the effectiveness of our MGAN against the baseline. 

\begin{table}[t]
	\centering
	\small
	\begin{tabular}{c|c|c}
		\hline
		      Method  & \textbf{R}   &  \textbf{HO}  \\ \hline\hline
		Baseline SPD (${L}_{0}$ loss in Eq.(\ref{loss:faster_rcnn})) &                   13.8      &     57.0      \\
		  Our MGAN (${L}_{0}$ + ${L}_{mask}$)  &                   11.9      &     52.7      \\
		  Our MGAN (${L}_{0}$ + $L_{occ}$)  &                   13.2      &   55.6      \\
		  Our Final MGAN (${L}_{0}$ + ${L}_{mask}$ + $L_{occ}$)   &        \textbf{11.5}      &     \textbf{51.7}     \\ \hline
	\end{tabular}
	\caption{Comparison (in log-average miss rates) of our MGAN with the baseline on the CityPersons. Best results are boldfaced. Beside our final MGAN (final row), we also show the performance of our MGA branch (second row) and occlusion-sensitive loss term (third row) alone. For fair comparison, we use the same training data, input scale ($\times1$) and network backbone (VGG-16). On the heavy occlusion set (\textbf{HO}), our detector significantly reduces the error from 57.0 to 51.7, compared to the baseline. }
	\label{tab:baseline_comparison} \vspace{-0.2cm}
\end{table}

\begin{table}[t]
	\centering
	\small
	\resizebox{\linewidth}{!}{
	\begin{tabular}{c|c|c}
		\hline
		      Set  & Dense Pixel-wise Annotations  &  Coarse-level Annotations \\ \hline\hline
		\textbf{R} &                   11.2      &    11.9     \\ \hline
		 \textbf{HO}  &                   51.7      &     52.7      \\
		   \hline
	\end{tabular}
	}

	\caption{Comparison (in log-average miss rates) of our MGAN detector when using dense pixel-wise labeling with coarse-level segmentation obtained through visible bounding box information in our MGA branch. Replacing former with latter in our MGA branch results in no significant deterioration in detection performance. On both sets, our approach based on coarse-level segmentation provides a trade-off between annotation cost and accuracy. }
	\label{tab:baseline_comparison_annotations} \vspace{-0.2cm}
\end{table}

\begin{table}[t]
	\centering
	\small

	\begin{tabular}{c|c|c|c}
		\hline
		      Set  & [50, 75]  & [75, 125] &  \textgreater 125 \\ \hline\hline
		Baseline SPD &                   66.3      & 59.7      &   43.1     \\ \hline
		Our MGAN  &                \textbf{61.7}          &   \textbf{52.3}      &   \textbf{37.6}      \\
		   \hline
	\end{tabular}

	\caption{Comparison (in log-average miss rates) by dividing pedestrians w.r.t. their height (pixels):
small [50-75], medium [75-125] and large (\textgreater125) representing 28\%, 37\%
and 35\%, respectively of CityPersons \textbf{HO} set. Best results are boldfaced in each case.}
	\label{tab:baseline_comparison_small} \vspace{-0.3cm}
\end{table}

\noindent\textbf{Comparison with other attention strategies.}
We compare our approach with attention strategies proposed by \cite{ShanshanCVPR18} . The work of \cite{ShanshanCVPR18}
investigates channel attention (CA), visible box attention
(CA-VBB) and part attention (CA-Part). Both CA and CA-VBB
exploit channel-wise attention, with the latter also using
VBB information. In addition, CA-Part utilizes a part detection
network pre-trained on MPII Pose dataset. In contrast to
CA-Part, our method does not require extra annotations for part detection. 

We perform an experiment integrating CA and CA-VBB attention strategies \cite{ShanshanCVPR18} in our framework. On the \textbf{R} and \textbf{HO} sets of CityPersons validation set, CA attention strategy achieives a log-average miss rate of $17.3\%$ and $54.5\%$, respectively. The CA-VBB attention scheme obtains a log-average miss rate of $14.0\%$ and $54.1\%$ on the \textbf{R} and \textbf{HO} sets, respectively. Our approach without ${L}_{occ}$ outperforms both CA and CA-VBB strategies on both \textbf{R} and \textbf{HO} sets by achieving a log-average miss rate of $11.9\%$ and $52.7\%$, respectively. 

\begin{table}[t]
	\centering
	\resizebox{\linewidth}{!}{
	\begin{tabular}{c|c|c|c|c|c|c}
		\hline
		              Method               & VBB & Backbone & Data (visibility) &    Scale    &  \textbf{R}   &  \textbf{HO}   \\ \hline\hline
		  OR-CNN \cite{Zhang_2018_ECCV}    &   $\checkmark$  & VGG & $\ge50\%$      &  $\times1$  &     12.8      &      55.7      \\
		  Our MGAN                &  $\checkmark$   & VGG  &   $\ge50 \%$        &  $\times1$  & \textbf{10.5} & \textbf{47.2}  \\ \hline
		   OR-CNN \cite{Zhang_2018_ECCV}    &   $\checkmark$  & VGG & $\ge50\%$      &  $\times1.3$  &     11.0      &      51.3      \\
		  Our MGAN                &  $\checkmark$   & VGG  &   $\ge50\%$        &  $\times1.3$  & \textbf{9.9} & \textbf{45.4}  \\ \hline
		  
		  ATT-vbb \cite{ShanshanCVPR18}   &  $\checkmark$  & VGG & $\ge65\%$      &  $\times1$  &     16.4      &      57.3      \\
		   Our MGAN                &  $\checkmark$   & VGG  &   $\ge65\%$        &  $\times1$  & \textbf{11.5} & \textbf{51.7}  \\ \hline
		 Bi-Box \cite{Zhou_2018_ECCV} &  $\checkmark$   & VGG &    $\ge30\%$        &  $\times1.3$  &     11.2      &      44.2      \\
		               Our MGAN                &  $\checkmark$   & VGG  &   $\ge30\%$        &  $\times1.3$  & \textbf{10.5}& \textbf{39.4}  \\ \hline\hline
	\end{tabular}}
	\caption{Comparison (in terms of log-average miss rate) with state-of-the-art methods that use both the visible bounding box (VBB)
	\textit{and} full body information on CityPersons validation set. For fair comparison, we use the same set of ground-truth pedestrian examples (visibility) and input scale for training when comparing with each method. Our MGAN outperforms all three methods on both sets. Under heavy occlusions (\textbf{HO}), our MGAN significantly reduces the error from 44.2 to 39.4, compared to the recently introduced Bi-Box \cite{Zhou_2018_ECCV}. Best results are boldfaced in each case.}
	\label{tab:citypersons_VBB_results}\vspace{-0.3cm}
\end{table}

\noindent\textbf{Impact of coarse-level segmentation.} As discussed in section \ref{subsubsection:Coarse-level Segmentation Annotation}, dense pixel-wise labelling is expensive to acquire. Further, such dense annotations are only available for CityPersons and not for Caltech dataset.  We validate our approach using coarse-level segmentation and compare it with using dense pixel-wise labelling in Tab.~\ref{tab:baseline_comparison_annotations}. On both sets, similar results are obtained with the coarse level information and dense pixel-wise labelling in our MGA branch. Our results in Tab.~\ref{tab:baseline_comparison_annotations} are also aligned to the prior work in instance segmentation \cite{Jifeng_Dai15}. Further, our final output is a detection box which does not require a precise segmentation mask prediction as in \cite{Jifeng_Dai15}. In addition, the difference between the two set of annotations is likely to reduce further for small pedestrians due to high-level of pooling operations undertaken by the network (\ie, we use RoI features from conv5\_3 of VGG). Our approach therefore provides a trade-off between annotation cost and accuracy.

\noindent\textbf{Heavy Occlusion and Size Variation.} We also evaluate the effectiveness of our approach on heavily occluded pedestrians with varying sizes, especially small pedestrians. Tab.~\ref{tab:baseline_comparison_small} shows that our approach provides improvement for all cases with a notable gain of 4.6\% for the small sized (50-75 pixels tall) heavily occluded pedestrians, compared to the baseline.

\begin{figure*}[t]
    \centering
    \resizebox{\linewidth}{!}{
    \begin{tabular}{cccc}
        \includegraphics[height=3.0cm]{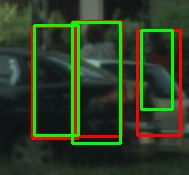}
         \includegraphics[height=3.0cm]{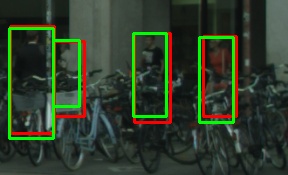}
         \includegraphics[height=3.0cm]{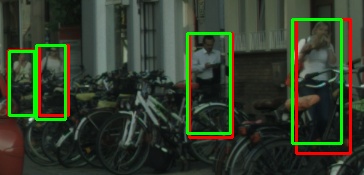}
    \end{tabular}
    
    }
    \caption{Detection examples on CityPersons dataset using our proposed pedestrian detector. The ground-truth and our detector predictions are shown in red and green respectively. Our detector accurately detects pedestrians under partial and heavy occlusions.}
    \label{fig:CityPersons_qual}\vspace{-0.2cm}
\end{figure*}

\subsection{State-of-the-art Comparison on CityPersons}
 
Our MGAN detector is compared to the recent state-of-the-art methods, namely Repulsion Loss \cite{XinlongWangCVPR18}, ATT-part \cite{ShanshanCVPR18}, ALFNet \cite{WeiLiuECCV18}, OR-CNN \cite{Zhang_2018_ECCV}, TLL \cite{Song_2018_ECCV}, Bi-Box \cite{Zhou_2018_ECCV} on CityPersons validation set. It is worth mentioning that existing pedestrian detection methods employ different set of ground-truth pedestrian examples for training. We therefore select  the same set of ground-truth pedestrian examples and input scale when comparing with each state-of-the-art method. Among existing methods, ATT-vbb \cite{ShanshanCVPR18}, OR-CNN \cite{Zhang_2018_ECCV} and Bi-Box \cite{Zhou_2018_ECCV} employ both the visible bounding box (VBB) and full body information similar to our method. We therefore first compare our approach with these three methods. Tab.~\ref{tab:citypersons_VBB_results} shows the the comparison in terms of log average miss rate (MR) on the \textbf{R}  and \textbf{HO} sets of CityPersons dataset. Our MGAN outperforms all three methods on both \textbf{R} and \textbf{HO} sets. When using an input scale of $1\times$, the OR-CNN method \cite{Zhang_2018_ECCV} employs both full body and visible region information and enforces the pedestrian proposals to be close and compactly located to corresponding objects, achieves a log-average miss rate of 12.8 and 55.7 on the \textbf{R} and \textbf{HO} sets, respectively. The detection results of OR-CNN \cite{Zhang_2018_ECCV} are improved when using an input scale of $1.3\times$. Our MGAN detector outperforms OR-CNN with a significant margin on both input scales. 

For an input scale of $1\times$, the ATT-vbb approach \cite{ShanshanCVPR18} employing FasterRCNN detector with a visible bounding box channel
attention net obtains a log-average miss rate 16.4 and 57.3 on the \textbf{R} and \textbf{HO} sets, respectively. Our MGAN provides superior detection results with a log-average miss rate of 11.5 and 51.7 on the \textbf{R} and \textbf{HO} sets, respectively. Moreover, the recently introduced Bi-Box method \cite{Zhou_2018_ECCV} utilizes visible bounding box (VBB) information to generate visible part regions for pedestrian proposal generation. On the \textbf{R} and \textbf{HO} sets, the Bi-Box approach \cite{Zhou_2018_ECCV} yields a log-average miss rate of 11.2 and 44.2, respectively using an input scale of $1.3\times$. Our MGAN outperforms Bi-Box on both sets by achieving a log-average miss rate of 10.5 and 39.4, respectively. 

To summarize, the results in Tab.~\ref{tab:citypersons_VBB_results} clearly signify the effectiveness of our MGAN towards handling heavy occlusions (\textbf{HO}) compared to these  methods \cite{ShanshanCVPR18,Zhang_2018_ECCV,Zhou_2018_ECCV} using \textit{same} level of supervision, ground-truth pedestrian examples during training, input scale and backbone. Tab.~\ref{tab:citypersons_results_all} further shows the comparison with all published state-of-the-art methods on the CityPersons. 
Fig.~\ref{fig:CityPersons_qual} displays example detections from our MGAN on CityPersons. Examples show a range of occlusion degrees \ie from partial to heavy. Finally, Tab.~\ref{tab:citypersons_test_results} shows the state-of-the-art comparison on the CityPersons test set.

\begin{table}[t]
	\centering
	\begin{tabular}{c|c|c|c|c}
		\hline
		              Method               & Data (visibility) &    Scale    &  \textbf{R}   &  \textbf{HO}   \\ \hline\hline
		    TLL \cite{Song_2018_ECCV}      &        -         &  $\times1$  &     14.4      &      52.0      \\ \hline
		  ATT-part \cite{ShanshanCVPR18}   &     $\ge65\%$      &  $\times1$  &     16.0      &      56.7      \\
		Rep. Loss \cite{XinlongWangCVPR18} &                  &  $\times1$  &     13.2      &      56.9      \\
		               MGAN                &                  &  $\times1$  & \textbf{11.5} & \textbf{51.7}  \\ \hline
		  OR-CNN \cite{Zhang_2018_ECCV}    &     $\ge50\%$      &  $\times1$  &     12.8      &      55.7      \\
		               MGAN                &                  &  $\times1$  & \textbf{10.5} & \textbf{47.2}  \\ \hline
		    ALF   \cite{WeiLiuECCV18}      &      $\ge0\%$      &  $\times1$  &     12.0      &     51.9      \\
		               MGAN                &                  &  $\times1$  &     \textbf{11.3}     &    \textbf{42.0}       \\ \hline
		Rep. Loss \cite{XinlongWangCVPR18} &     $\ge65\%$      & $\times1.3$ &    {11.6}     &      55.3      \\
		               MGAN                &                  & $\times1.3$ & \textbf{10.3} & \textbf{49.6}  \\ \hline
		  OR-CNN \cite{Zhang_2018_ECCV}    &     $\ge50\%$      & $\times1.3$ &     11.0      &      51.3      \\
		               MGAN                &                  & $\times1.3$ & \textbf{9.9} & \textbf{45.4 } \\ \hline
		   Bi-Box \cite{Zhou_2018_ECCV}    &     $\ge30\%$      & $\times1.3$ &     11.2      &      44.2      \\
		               MGAN                &                  & $\times1.3$ & \textbf{10.5} & \textbf{39.4 } \\ \hline\hline
	\end{tabular}
	\caption{Comparison (in terms of log-average miss rate) of MGAN with state-of-the-art methods in literature on CityPersons validation set. Our MGAN sets a new state-of-the-art by outperforming all existing methods. Best results are boldfaced in each case.}
	\label{tab:citypersons_results_all}\vspace{-0.2cm}
\end{table}

\begin{table}[t]
	\centering
		\begin{tabular}{c|c|c}
		
			\hline
			                  Method                    &  \textbf{R}   &  \textbf{HO}   \\ \hline\hline
			Adaptive Faster RCNN \cite{citypersons_2017_zhang} &     12.97     &     50.47      \\
			       Rep. Loss \cite{XinlongWangCVPR18}        &     11.48     &     52.59      \\
			       OR-CNN \cite{Zhang_2018_ECCV}        &     11.32     &     51.43      \\
			                 Our MGAN                   & \textbf{9.29} & \textbf{40.97} \\ \hline\hline
		\end{tabular}
	\caption{Comparison (in terms of log-average miss rate) of MGAN with state-of-the-art methods on CityPersons test set. The test set is withheld and results are obtained by sending our detection predictions to the authors of CityPersons dataset \cite{citypersons_2017_zhang} for evaluation.}
	\label{tab:citypersons_test_results}\vspace{-0.2cm}
\end{table}

\begin{figure*}[t]
    \centering
    \resizebox{\linewidth}{!}
     {
    \subfigure[\textbf{R}]{\includegraphics[width = 0.33\linewidth]{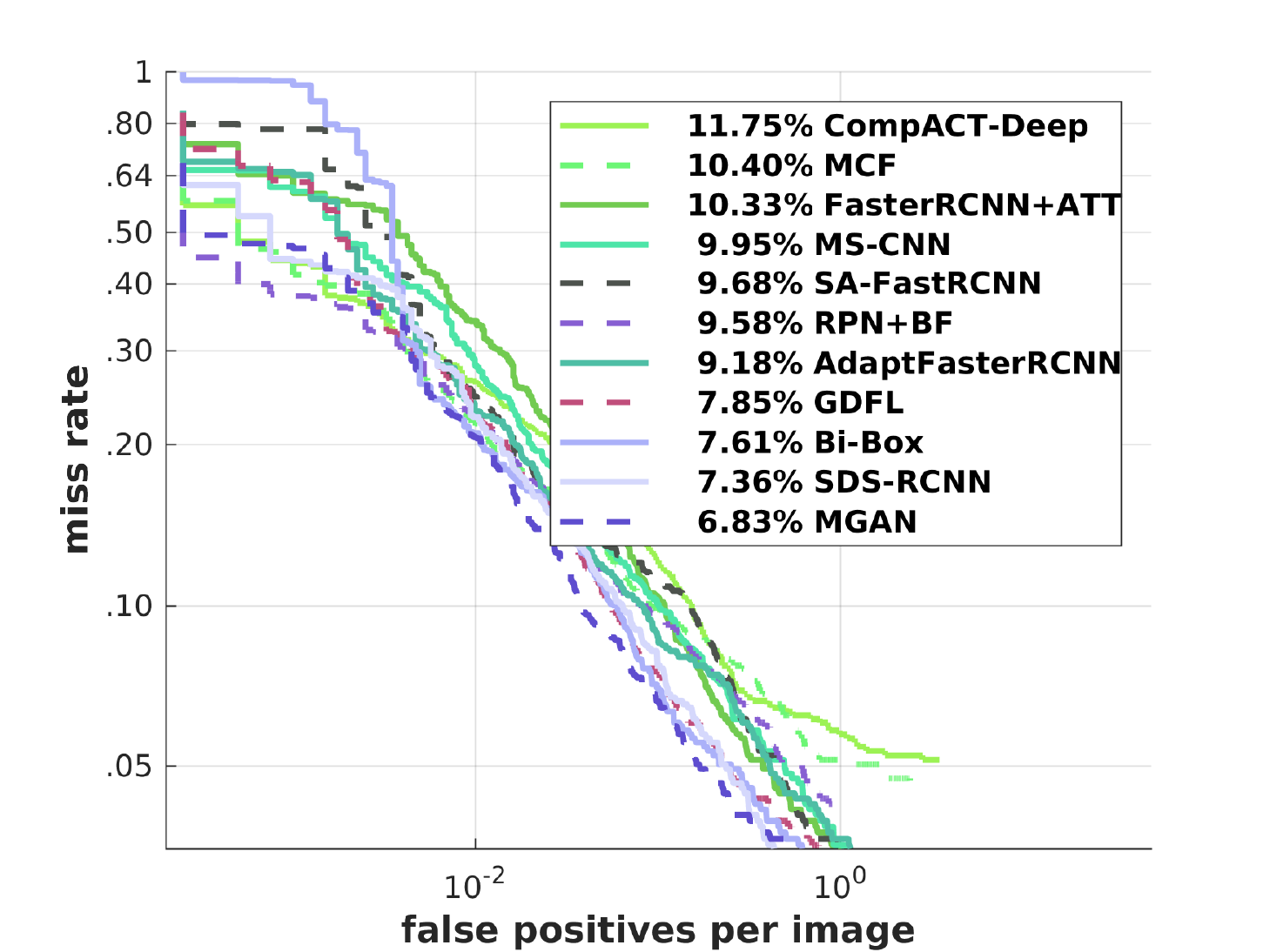}}
    \subfigure[\textbf{HO}]{\includegraphics[width = 0.33\linewidth]{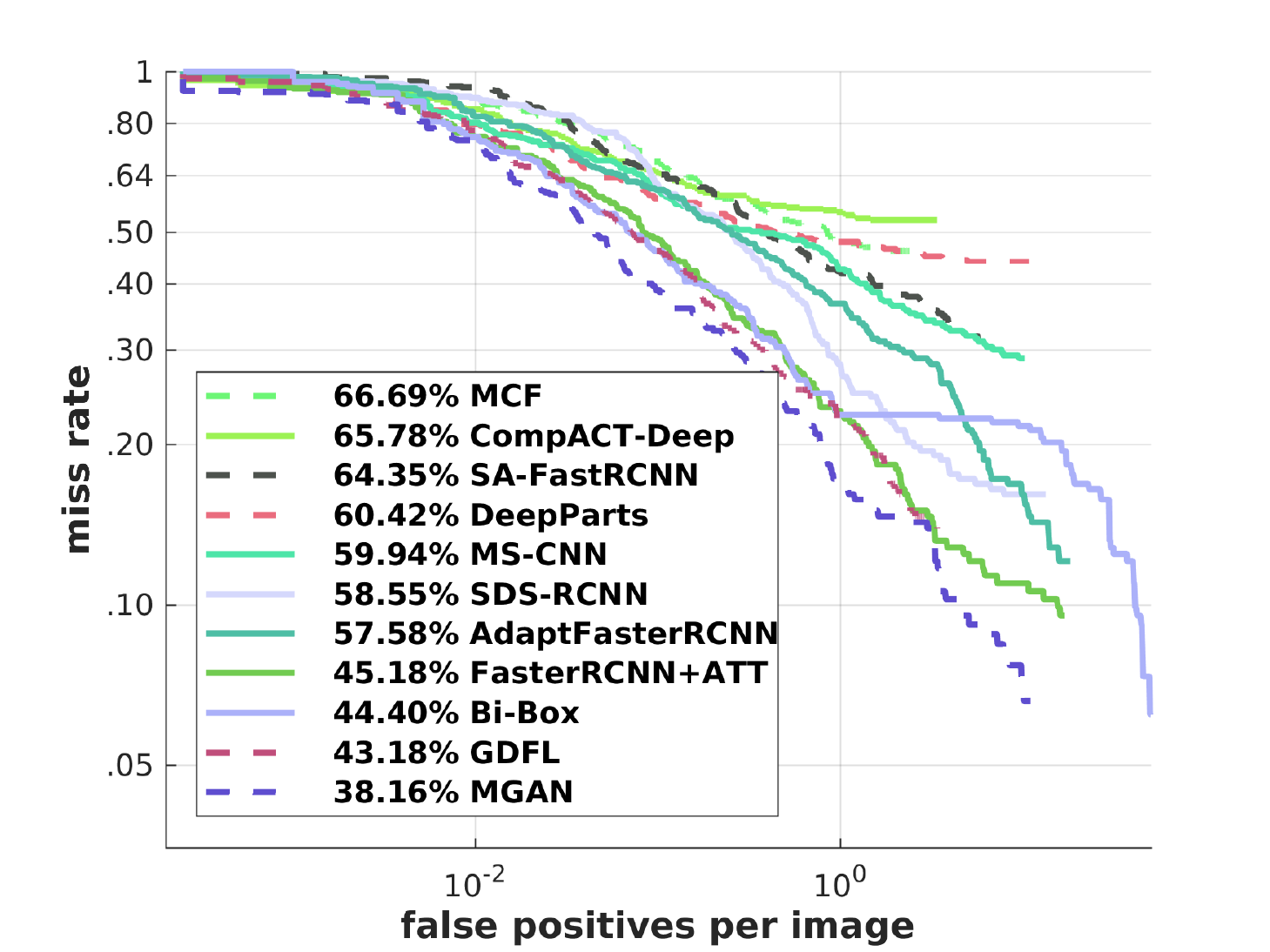}}
    \subfigure[\textbf{R+HO}]{\includegraphics[width = 0.33\linewidth]{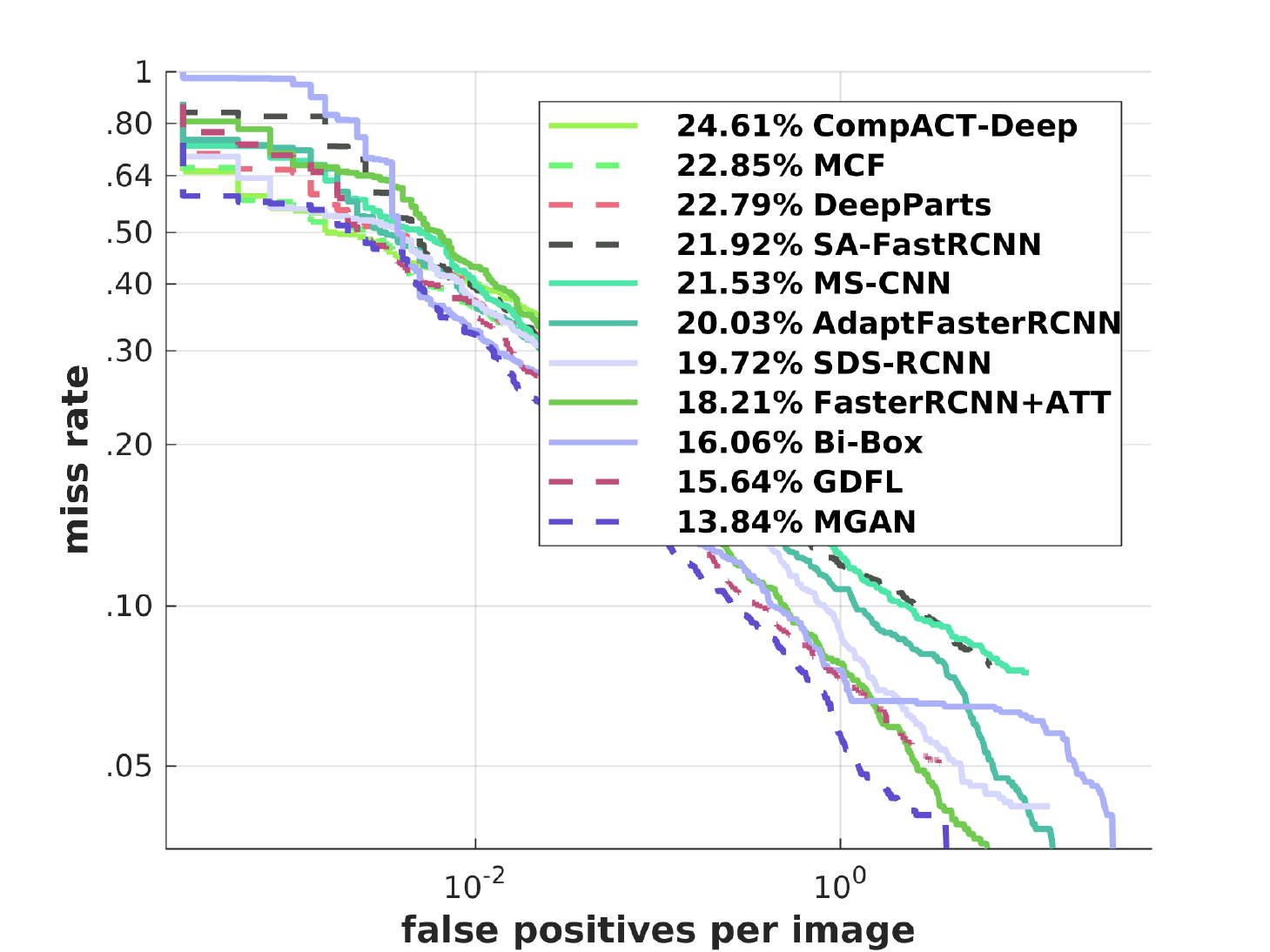}}}   
    
    \caption{State-of-the-art comparison on the \textbf{R}, \textbf{HO} and \textbf{R+HO} subsets of Caltech dataset. The legend in each plot represents the log-averaged miss rate over FPPI=$[{10}^{-2},{10}^{0}]$. Our approach provides superior results compared to existing approaches on all three subsets. }
    \label{fig:caltech-result-graph} \vspace{-0.4cm}
\end{figure*}

\subsection{Caltech Dataset} Here, MGAN is compared to the following recent state-of-art methods: CompACT-Deep \cite{Cai_2015_ICCV}, DeepParts\cite{YonglongTianICCV15}, MS-CNN \cite{ZhaoweiCaiECCV16}, RPN+BF \cite{zhang2016faster}, SA-F.RCNN \cite{li2018scale}, MCF \cite{cao2017MCF}, SDS-RCNN \cite{GarrickBrazilICCV17}, F.RCNN \cite{citypersons_2017_zhang}, F.RCNN+ATT-vbb \cite{ShanshanCVPR18},  GDFL \cite{Lin_2018_ECCV}, and  Bi-Box \cite{Zhou_2018_ECCV}. 
Tab.~\ref{tab:caltech_results} compares MGAN with the state-of-the-art methods under all three occlusion subsets:  \textbf{R},  \textbf{HO} and   \textbf{R+HO}. Among existing methods, the SDS-RCNN approach \cite{GarrickBrazilICCV17} reports a log-average miss rate of 7.36 on the \textbf{R} set. Our MGAN achieives superior results with a log-average miss rate of 6.83 on this set. On the \textbf{HO} and \textbf{R+HO} sets, the GDFL detector \cite{Lin_2018_ECCV} provides the best results among the existing methods with a log-average miss rate of 43.18 and 15.64, respectively. Our MGAN detector outperforms GDFL with an absolute gain of 5.02\% and 1.80\% on \textbf{HO} and \textbf{R+HO} sets, respectively. Fig.~\ref{fig:caltech-result-graph} shows the comparison of our detector with existing methods over the whole spectrum of false positives per image metric.

\begin{table}[t]
	\resizebox{\linewidth}{!}{
		\begin{tabular}{c|c|c|c|c}
			\hline
			              Detector               &   Occl.    &  \textbf{R}   &  \textbf{HO}   & \textbf{R+HO}  \\ \hline\hline
			 CompACT-Deep \cite{Cai_2015_ICCV}   &  $\times$  &     11.75     &     65.78      &     24.61      \\
			 DeepParts\cite{YonglongTianICCV15}  & \checkmark &     11.89     &     60.42      &     22.79      \\
			 MCF \cite{cao2017MCF}               &  $\times$  &     10.40      &     66.69      &     22.85      \\
			   ATT-part \cite{ShanshanCVPR18}    & \checkmark &     10.33     &     45.18      &     18.21      \\
			   MS-CNN \cite{ZhaoweiCaiECCV16}    &  $\times$  &     9.95      &     59.94      &     21.53      \\
			   RPN+BF \cite{zhang2016faster}     &  $\times$  &     9.58      &     74.36      &     24.01      \\
			    SA-F.RCNN \cite{li2018scale}     &  $\times$  &     9.68      &     64.35      &     21.92      \\
			SDS-RCNN \cite{GarrickBrazilICCV17}  &  $\times$  &     7.36      &     58.55      &     19.72      \\
			F.RCNN \cite{citypersons_2017_zhang} &  $\times$  &     9.18      &     57.58      &     20.03      \\
			     GDFL \cite{Lin_2018_ECCV}       &  $\times$  &     7.85      &     43.18      &     15.64      \\
			    Bi-Box \cite{Zhou_2018_ECCV}     & \checkmark &     7.61      &     44.40      &     16.06      \\ \hline
			              Our  MGAN                 & \checkmark & \textbf{6.83} & \textbf{38.16} & \textbf{13.84} \\ \hline\hline
		\end{tabular}
	}
	\caption{Comparison (in terms of log-average miss rate) of MGAN with the state-of-art methods on the Caltech dataset. The second column indicates whether the method is specifically targeted to handling occlusion. Best results are in bold. Under heavy occlusions (\textbf{HO}), our detector outperforms the state-of-the-art GDFL detector by 5.0\%. Further, our detector provides superior results compared to all published methods on both the reasonable (\textbf{R}) and the combined set of reasonable and heavy occlusions (\textbf{R+HO}). } \vspace{-0.4cm}
	\label{tab:caltech_results}
\end{table}

\begin{figure}[t]
    \centering
        \resizebox{\linewidth}{!}{
        \begin{tabular}{ccccc}
        \includegraphics[height=3cm]{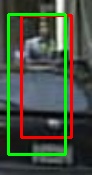}& 
        \includegraphics[height=3cm]{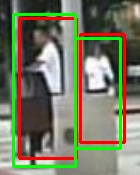}&
        \includegraphics[height=3cm]{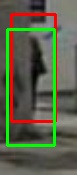}&
        \includegraphics[height=3cm]{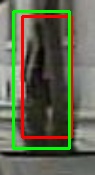}&
        \includegraphics[height=3cm]{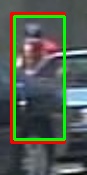}\\
        \multicolumn{5}{c}{(a) MGAN} \\
        \includegraphics[height=3cm]{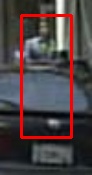}& 
        \includegraphics[height=3cm]{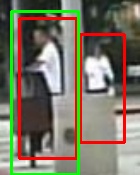}& 
        \includegraphics[height=3cm]{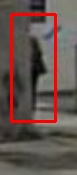}&
        \includegraphics[height=3cm]{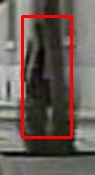}&
         \includegraphics[height=3cm]{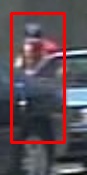}\\
        \multicolumn{5}{c}{(b) ATT-vbb \cite{ShanshanCVPR18}}\\
        \includegraphics[height=3cm]{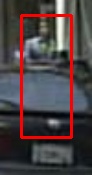}&
        \includegraphics[height=3cm]{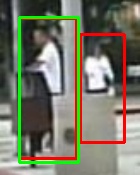}&        
        \includegraphics[height=3cm]{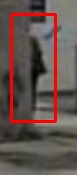}&
        \includegraphics[height=3cm]{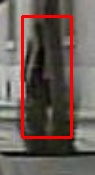}&
         \includegraphics[height=3cm]{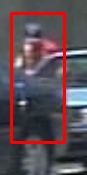}\\
         \multicolumn{5}{c}{(c) GDFL \cite{Lin_2018_ECCV}}\\
        \end{tabular}
        }

    \caption{Qualitative comparison of (a) MGAN with (b) ATT-vbb \cite{ShanshanCVPR18} and (c) GDFL \cite{Lin_2018_ECCV} on Caltech test set. Red boxes denote the ground-truth and green boxes indicate detector predictions. Examples images depict varying level of occlusions.}
    \label{fig:vis_detectors_caltech} \vspace{-0.3cm}
\end{figure}

We further signify the effectiveness of MGAN towards handling occlusions by drawing visual comparison with ATT-vbb \cite{ShanshanCVPR18}, and GDFL \cite{Lin_2018_ECCV} in Fig.~\ref{fig:vis_detectors_caltech}. All results are obtained using the same FPPI. Our MGAN accurately  detects pedestrians in all five scenarios.

\section{Conclusion}
We proposed a mask-guided attention network (MGAN) for occluded pedestrian detection. The MGA module generates spatial attention mask using visible body region information. The resulting spatial attention mask modulates the full body features (\ie, highlighting the features of pedestrian visible region, and suppressing the background). Instead of dense pixel labelling, we employ coarse-level segmentation information for visible region. In addition to MGA, we introduced an occlusion-sensitive loss term. Experiments on two datasets clearly show the effectiveness of our approach, especially for heavily occluded pedestrians.

\noindent\textbf{Acknowledgments} This work was supported by National Natural Science Foundation of China (Grant \# 61632018).

{\small
\bibliographystyle{ieee_fullname}
\bibliography{egbib}

\begin{thebibliography}{10}\itemsep=-1pt

\bibitem{GarrickBrazilICCV17}
Garrick Brazil, Xi Yin, and Xiaoming Liu.
\newblock Illuminating pedestrians via simultaneous detection \& segmentation.
\newblock In {\em ICCV}, 2017.

\bibitem{ZhaoweiCaiECCV16}
Zhaowei Cai, Quanfu Fan, Rogerio~Schmidt Feris, and Nuno Vasconcelos.
\newblock A unified multi-scale deep convolutional neural network for fast
  object detection.
\newblock In {\em ECCV}, 2016.

\bibitem{Cai_2015_ICCV}
Zhaowei Cai, Mohammad Saberian, and Nuno Vasconcelos.
\newblock Learning complexity-aware cascades for deep pedestrian detection.
\newblock In {\em ICCV}, 2015.

\bibitem{cao2017MCF}
Jiale Cao, Yanwei Pang, and Xuelong Li.
\newblock Learning multilayer channel features for pedestrian detection.
\newblock {\em TIP}, 26(7):3210--3220, July 2017.

\bibitem{Di_Cheneccv18}
Di Chen, Shanshan Zhang, Wanli Ouyang, Jian Yang1, and Ying Tai.
\newblock Person search via a mask-guided two-stream cnn model.
\newblock In {\em ECCV}, 2018.

\bibitem{Jifeng_Dai15}
Jifeng Dai, Kaiming He, and Jian Sun.
\newblock Boxsup: Exploiting bounding boxes to supervise convolutional networks
  for semantic segmentation.
\newblock In {\em ICCV}, 2015.

\bibitem{Dollar_2012_PAMI}
Piotr Doll{\'{a}}r, Christian Wojek, Bernt Schiele, and Pietro Perona.
\newblock Pedestrian detection: An evaluation of the state of the art.
\newblock {\em TPAMI}, 34(4):743--761, April 2012.

\bibitem{du2017fused}
Xianzhi Du, Mostafa El-Khamy, Jungwon Lee, and Larry Davis.
\newblock Fused dnn: A deep neural network fusion approach to fast and robust
  pedestrian detection.
\newblock In {\em WACV}, 2017.

\bibitem{maskrcnn_2017_iccv_he}
Kaiming He, Georgia Gkioxari, Piotr Doll{\'a}r, and Ross Girshick.
\newblock Mask r-cnn.
\newblock In {\em ICCV}, 2017.

\bibitem{He_2016_CVPR}
Kaiming He, Xiangyu Zhang, Shaoqing Ren, and Jian Sun.
\newblock Deep residual learning for image recognition.
\newblock In {\em CVPR}, 2016.

\bibitem{kingma2014adam}
Diederik~P Kingma and Jimmy Ba.
\newblock Adam: A method for stochastic optimization.
\newblock {\em arXiv preprint arXiv:1412.6980}, 2014.

\bibitem{li2018scale}
Jianan Li, Xiaodan Liang, ShengMei Shen, Tingfa Xu, Jiashi Feng, and Shuicheng
  Yan.
\newblock Scale-aware fast r-cnn for pedestrian detection.
\newblock {\em TMM}, 20(4):985--996, April 2018.

\bibitem{Lin_2018_ECCV}
Chunze Lin, Jiwen Lu, Gang Wang, and Jie Zhou.
\newblock Graininess-aware deep feature learning for pedestrian detection.
\newblock In {\em ECCV}, 2018.

\bibitem{lin2018focal}
Tsung-Yi Lin, Priyal Goyal, Ross Girshick, Kaiming He, and Piotr Doll{\'a}r.
\newblock Focal loss for dense object detection.
\newblock {\em TPAMI}, 2018.

\bibitem{liu2016ssd}
Wei Liu, Dragomir Anguelov, Dumitru Erhan, Christian Szegedy, Scott Reed,
  Cheng-Yang Fu, and Alexander~C Berg.
\newblock Ssd: Single shot multibox detector.
\newblock In {\em ECCV}, 2016.

\bibitem{WeiLiuECCV18}
Wei Liu, Shengcai Liao, Weidong Hu, Xuezhi Liang, and Xiao Chen.
\newblock Learning efficient single-stage pedestrian detectors by asymptotic
  localization fitting.
\newblock In {\em ECCV}, 2018.

\bibitem{JiayuanMaoCVPR17}
Jiayuan Mao, Tete Xiao, Yuning Jiang, and Zhimin Cao.
\newblock What can help pedestrian detection?
\newblock In {\em CVPR}, 2017.

\bibitem{MarkusMathiasICCV13}
Markus Mathias, Rodrigo Benenson, Radu Timofte, and Luc~Van Gool.
\newblock Handling occlusions with franken-classifiers.
\newblock In {\em ICCV}, 2013.

\bibitem{JunhyugNohCVPR18}
Junhyug Noh, Soochan Lee, Beomsu Kim, and Gunhee Kim.
\newblock Improving occlusion and hard negative handling for single-stage
  pedestrian detectors.
\newblock In {\em CVPR}, 2018.

\bibitem{WanliOuyangICCV13}
Wanli Ouyang and Xiaogang Wang.
\newblock Joint deep learning for pedestrian detection.
\newblock In {\em ICCV}, 2013.

\bibitem{cascade}
Yanwei Pang, Jiale Cao, and Xuelong Li.
\newblock Cascade learning by optimally partitioning.
\newblock {\em TCyb}, 47(12):4148--4161, Dec 2017.

\bibitem{CIC}
Yanwei Pang, Manli Sun, Xiaoheng Jiang, and Xuelong Li.
\newblock Convolution in convolution for network in network.
\newblock {\em TNNLS}, 29(5):1587--1597, May 2018.

\bibitem{JimmyRenCVPR17}
Jimmy Ren, Xiaohao Chen, Jianbo Liu, Wenxiu Sun, Jiahao Pang, Qiong Yan,
  Yu-Wing Tai, and Li Xu.
\newblock Accurate single stage detector using recurrent rolling convolution.
\newblock In {\em CVPR}, 2017.

\bibitem{fasterrcnn_2015_nips}
Shaoqing Ren, Kaiming He, Ross Girshick, and Jian Sun.
\newblock Faster r-cnn: Towards real-time object detection with region proposal
  networks.
\newblock In {\em NIPS}, 2015.

\bibitem{simonyan2014vgg}
Karen Simonyan and Andrew Zisserman.
\newblock Very deep convolutional networks for large-scale image recognition.
\newblock {\em arXiv preprint arXiv:1409.1556}, 2014.

\bibitem{Song_2018_ECCV}
Tao Song, Leiyu Sun, Di Xie, Haiming Sun, and Shiliang Pu.
\newblock Small-scale pedestrian detection based on topological line
  localization and temporal feature aggregation.
\newblock In {\em ECCV}, 2018.

\bibitem{GlanceNets}
Hanqing Sun and Yanwei Pang.
\newblock Glancenets — efficient convolutional neural networks with adaptive
  hard example mining.
\newblock {\em Science China Information Sciences}, 61(10):101--109, Oct 2018.

\bibitem{YonglongTianICCV15}
Yonglong Tian, Ping Luo, Xiaogang Wang, and Xiaoou Tang.
\newblock Deep learning strong parts for pedestrian detection.
\newblock In {\em ICCV}, 2015.

\bibitem{XinlongWangCVPR18}
Xinlong Wang, Tete Xiao, Yuning Jiang, Shuai Shao, Jian Sun, and Chunhua Shen.
\newblock Repulsion loss: Detecting pedestrians in a crowd.
\newblock In {\em CVPR}, 2018.

\bibitem{zhang2016faster}
Liliang Zhang, Liang Lin, Xiaodan Liang, and Kaiming He.
\newblock Is faster r-cnn doing well for pedestrian detection?
\newblock In {\em ECCV}, 2016.

\bibitem{citypersons_2017_zhang}
Shanshan Zhang, Rodrigo Benenson, and Bernt Schiele.
\newblock Citypersons: A diverse dataset for pedestrian detection.
\newblock In {\em CVPR}, 2017.

\bibitem{Zhang_2018_ECCV}
Shifeng Zhang, Longyin Wen, Xiao Bian, Zhen Lei, and Stan~Z. Li.
\newblock Occlusion-aware r-cnn: Detecting pedestrians in a crowd.
\newblock In {\em ECCV}, 2018.

\bibitem{ShanshanCVPR18}
Shanshan Zhang, Jian Yang, and Bernt Schiele.
\newblock Occluded pedestrian detection through guided attention in cnns.
\newblock In {\em CVPR}, 2018.

\bibitem{zhou2014non}
Chunluan Zhou and Junsong Yuan.
\newblock Non-rectangular part discovery for object detection.
\newblock In {\em BMVC}, 2014.

\bibitem{ChunluanZhouICCV17}
Chunluan Zhou and Junsong Yuan.
\newblock Multi-label learning of part detectors for heavily occluded
  pedestrian detection.
\newblock In {\em ICCV}, 2017.

\bibitem{Zhou_2018_ECCV}
Chunluan Zhou and Junsong Yuan.
\newblock Bi-box regression for pedestrian detection and occlusion estimation.
\newblock In {\em ECCV}, 2018.

\end{thebibliography}
}

\end{document}